%% file: main.tex
\documentclass[10pt,twocolumn,letterpaper]{article}

\usepackage{cvpr}              

\input{preamble}

\usepackage[pagebackref,breaklinks,colorlinks,citecolor=cvprblue]{hyperref}

\usepackage{url}

\usepackage{graphicx}
\usepackage{booktabs}       
\usepackage{amsfonts}       
\usepackage{nicefrac}       
\usepackage{microtype}      
\usepackage{xcolor}         
\usepackage{amsmath}
\usepackage{enumitem}
\usepackage{subcaption}
\usepackage{multirow}
\usepackage[capitalize,noabbrev]{cleveref}
\usepackage[accsupp]{axessibility}
\graphicspath{{figures/}} 

\definecolor{cvprblue}{rgb}{0.21,0.49,0.74}

\def\paperID{46} 
\def\confName{CVPR}
\def\confYear{2024}

\title{Benchmarking Zero-Shot Recognition with Vision-Language Models:\\ Challenges on Granularity and Specificity}

\author{ Zhenlin Xu\textsuperscript{1}\thanks{Correspondence to: \texttt{xzhenlin@amazon.com}} $\quad$ Yi Zhu$^2$\thanks{Work done while at Amazon} $\quad$ Siqi Deng$^1$ $\quad$ Abhay Mittal\textsuperscript{3\textdagger} $\quad$ Yanbei Chen$^1$ $\quad$ Manchen Wang\textsuperscript{3\textdagger}  $\quad$ \\
{Paolo Favaro$^1$ $\quad$ Joseph Tighe\textsuperscript{3\textdagger} $\quad$ Davide Modolo$^1$}  \\
$^1$AWS AI Labs $\quad$ $^2$Boson AI $\quad$ $^3$Meta\\
}

\begin{document}
\maketitle

\begin{abstract}
This paper presents novel benchmarks for evaluating vision-language models (VLMs) in zero-shot recognition, focusing on \textit{granularity} and \textit{specificity}. Although VLMs excel in tasks like image captioning, they face challenges in open-world settings. Our benchmarks test VLMs' consistency in understanding concepts across semantic granularity levels and their response to varying text specificity. Findings show that VLMs favor moderately fine-grained concepts and struggle with specificity, often misjudging texts that differ from their training data. Extensive evaluations reveal limitations in current VLMs, particularly in distinguishing between correct and subtly incorrect descriptions. While fine-tuning offers some improvements, it doesn't fully address these issues, highlighting the need for VLMs with enhanced generalization capabilities for real-world applications. This study provides insights into VLM limitations and suggests directions for developing more robust models.
\end{abstract}

\section{Introduction}
Vision-language models (VLMs) have shown impressive capabilities in a wide range of tasks, including image captioning~\cite{wang2022beitv3,yu2022CoCa}, visual question answering~\cite{chen2023pali}, and notably, zero-shot visual recognition~\cite{radford2021clip,zhou2022detecting,gu2021vild,xu2022groupvit}. Models pretrained on large-scale image-caption datasets ~\cite{jia2021align,yuan2021florence,Schuhmann2021LAION400MOD,Schuhmann2022LAION5B}, like CLIP~\cite{radford2021clip}, have been at the forefront. These models achieve this by mapping visual and linguistic inputs to a shared latent space, enabling the recognition of novel objects or concepts in a zero-shot manner—a capability critical for developing versatile and intelligent visual systems.

While current VLMs perform well in various tasks, their application in open-world scenarios poses unique challenges. An ideal open-world zero-shot model would recognize any language-defined input, from simple concepts like ``an image of flowers" to more complex descriptions like ``a person playing with a dog on the beach", and output scores indicating whether the visual input \textit{correctly} implies the semantics of the language input. Existing works often evaluate the zero-shot capability on various classification dataset like ImageNet~\cite{russakovsky2015imagenet} and domain specific datasets \cite{li2022elevater} without the notion of granularity of concepts, as well as image and text retrieval on Flickr30K~\cite{plummer2015flickr30k} and COCO~\cite{lin2014microsoft} that are not able to reveal the general failure pattern. These benchmark fall short of replicating the complexities of a realistic open-world setting, leaving a substantial gap in our understanding of the effectiveness of VLMs in such scenarios.

\begin{figure*}[tp]
     \centering
     \begin{minipage}{0.49\textwidth}
        \includegraphics[width=\linewidth]{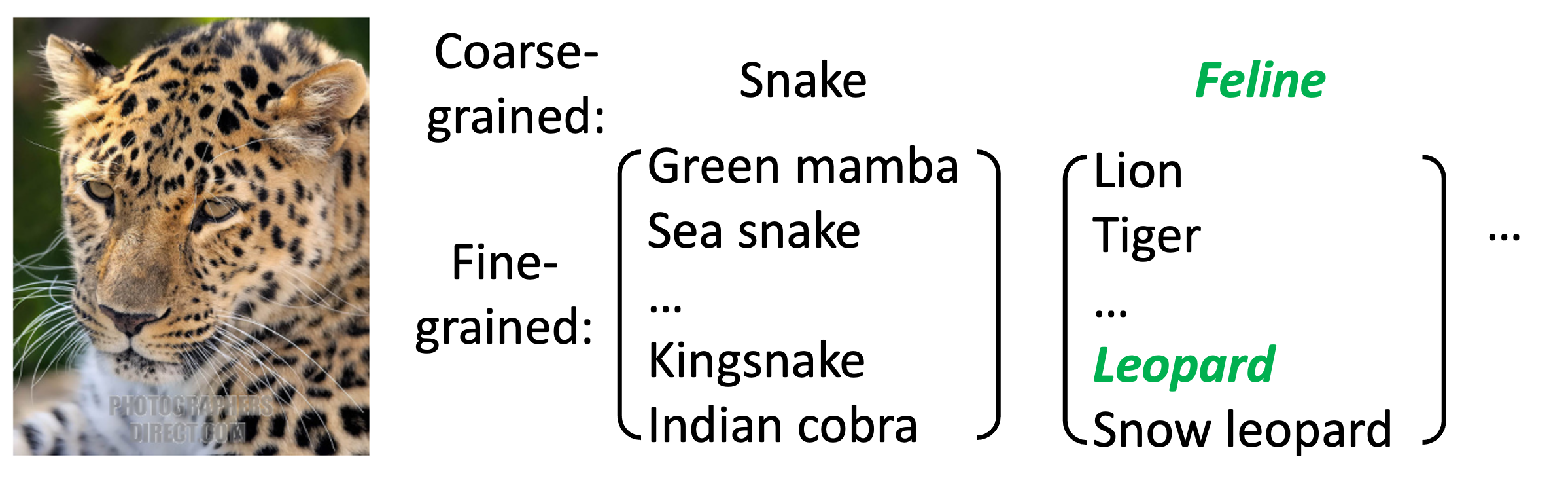}
        \end{minipage}
        \begin{minipage}{0.49\textwidth}
        \includegraphics[width=\linewidth]{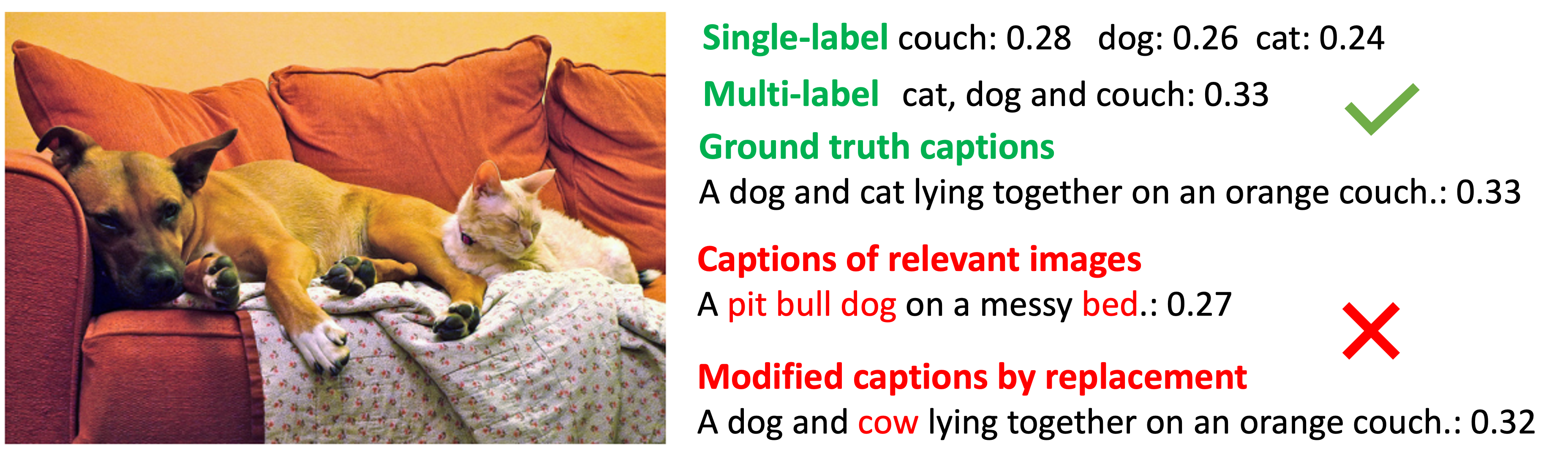}
        \end{minipage}
\caption{\textbf{Left}: Zero-shot models should recognize images with fine-grained (FG) concepts such as ``Leopard'', as well as coarse-grained (CG) concepts like ``Feline'' However, they often exhibit performance discrepancies on concepts at different levels of granularity.
\textbf{Right}: Zero-shot models should recognize whether the text correctly describe the given image. However, vision-language models could be sensitive to the specificity of text and struggle to distinguish between the challenging positive like single-label prompts and hard negatives like poisoned captions with small changes.}
\label{fig:paper_concept}
\end{figure*}

This paper present new benchmarks on the pivotal properties when deploying VLMs for real-world zero-shot recognition: \textit{granularity} and \textit{specificity}. The first benchmark examines VLMs' ability to consistently understand concepts at different levels of \textit{granularity}. For instance, a model should recognize an image of a leopard both when presented with a fine-grained query like “an image of a leopard” and a more coarse-grained query such as “an image of a feline.” This consistency is crucial, not just as an indicator of the model's comprehension of concept relationships but also for practical applications. A pertinent example is in autonomous driving, where recognizing “road cone” but failing to identify “barrier” could be problematic. To assess this, we employ an evaluation protocol where we measure the performance discrepancy in recognizing a coarse-grained class, both by directly using the coarse-grained class prompt and by aggregating predictions from its fine-grained children classes. We leverage a dataset with hierarchical labels, adapting ImageNet and its semantic hierarchy from WordNet.

The second benchmark evaluates how the specificity of language inputs affects VLM outputs, even when visual and linguistic inputs align. For example, a simple prompt like “a picture with a dog” may receive a lower score compared to a more detailed but incorrect caption “a dog and cow lying together on an orange couch.” This distinction is key in determining whether VLMs can accurately reflect the correctness of the alignment between visual and language inputs, rather than just overall similarity. To test this, we use an image-to-text retrieval task on the MS-COCO dataset, designing challenging positive texts with varying levels of specificity, such as single-label prompts with limited information, and hard negative texts like slightly modified but incorrect captions.

Our carefully designed benchmarks led to an extensive evaluation of state-of-the-art vision-language models (VLMs). We focus on contrastive models like CLIP, covering various aspects, including pretraining datasets, architectural designs, cross-modality interactions, and learning objectives. We discovered that VLMs face significant challenges in both benchmarks.

In the granularity evaluation, we find that \textit{VLMs prefer moderately fine-grained concepts over more abstract, coarse-grained ones.} This tendency seems closely tied to the nature of the training data. A detailed analysis of the LAION dataset revealed a higher presence of moderately fine-grained concepts in image alt-text, suggesting that data distribution plays a critical role.
In the specificity evaluation, VLMs showed \textit{sensitivity to text specificity}: texts that differ in specificity from the training data, such as straightforward single-label prompts or overly detailed captions, often received lower scores than more precisely detailed but slightly erroneous captions. This challenges the VLMs' ability to accurately distinguish between correct and subtly incorrect descriptions, complicating the retrieval of hard positive texts from hard negatives. The implication is that VLMs' scoring does not always reliably reflect the true alignment between visual and textual inputs. Our exploration into fine-tuning VLMs with these hard text samples revealed that while it offers some improvements, it does not fully resolve the challenges as a complete solution.

To our best knowledge, this is the first comprehensive study that evaluating VLMs from the perspective of semantic granularity and specificity. We believe that the carefully designed benchmark provide a valuable tool to the community to better quantitatively evaluate VLMs. With the proposed benchmark, we observed that all models surprisingly performs significantly worse than what we may hope. The findings and insights from our analysis may shed lights on better understanding the limitations of current VLMs and the challenges of using it for zero-shot recognition, and inspire new models with better generalization.

\section{Related Works}

\paragraph{Zero-shot visual recognition} 
CLIP-like vision-language foundation models have enabled open vocabulary visual recognition by mapping images with their corresponding language descriptions.
Early methods~\cite{radford2021clip,jia2021align} demonstrate the effectiveness of this paradigm on the image classification tasks.
For example, CLIP is able to achieve decent zero-shot accuracy on 27 image classification datasets. 
Given its potential, the language-driven visual recognition paradigm has been extended to tasks including object detection~\cite{zhou2022detecting}, semantic segmentation~\cite{xu2022groupvit}, video action recognition~\cite{wang2021actionclip}, depth estimation~\cite{zhang2022depthclip}, etc. 
Such language guided visual recognition has become the new paradigm in the field of computer vision since it can recognition new objects without any training data. 
In this paper, we would like to stress test these VLMs in terms of zero-shot visual recognition to better understand their capability and limitation in realistic open-world setting.

\paragraph{Benchmarking vision-language models} 
Thanks to the larger datasets and larger transformer models, many powerful vision-language models have been developed and shown great capability~\cite{yu2022CoCa,wang2022beitv3,chen2023pali}. 
At the same time, these models are being studied from various perspectives, such as robustness, bias, and other limitations~\cite{UnderstandingCLIP,StanislavPixels,goh2021multimodal,Noever2021ReadingIB,daras2022discovering}. 
~\cite{qiu2022aremm} investigates the robustness of nine open-sourced image-text models under common perturbations on five tasks, while ~\cite{schiappa2022robustness} studies the robustness of video-text models. 
~\cite{fang2022dataclip} further analyzes the robustness of VLMs under challenging natural distribution shifts and show that the more diverse training distribution is the main cause for the robustness gains.
~\cite{yuksekgonul2022vlmbow, thrush2022winoground} systematically evaluates the ability to encode compositional information of the VLMs.
~\cite{Cho2022DallEval} investigates the visual reasoning capabilities and social biases of different text-to-image models.
To improve transferability, ~\cite{shen2022klite} designs an efficient and scalable approach that leverages external knowledge to learn image representations.
In this paper, we study VLMs from two new perspectives: granularity and specificity through the lens of zero-shot visual recognition.


\begin{table}[htp]
\caption{An overview of the differences between the vision-language models evaluated in our study by the architecture, pretraining datasets, learning objectives, and if using cross-modality fusion modules. ITC, ITM, MIM, MTM, MMM stands for image-text contrastive, image-text matchng, masked image modeling, masked text modeling and masked multimodal modeling losses.}
\label{table:models}
\begin{center}
\begin{scriptsize}
\begin{tabular}{ccccc}
\toprule
Model & Architecture &  Datasets & Objectives &  Fusion\\
\midrule 
 \multirow{2}{*}{CLIP}  
 &   ViT-B-32 &  \multirow{2}{*}{\tiny Private400M}     &  \multirow{2}{*}{ITC} & \multirow{2}{*}{- }  \\
 & ViT-L-14 & & & \\
\midrule
 \multirow{4}{*}{OpenCLIP}  
&  ViT-B-32 &  \tiny LAION400M & \multirow{4}{*}{ITC}  & \multirow{4}{*}{-}   \\
  \cmidrule{2-3}
& ViT-B-32 &  \multirow{3}{*}{\tiny LAION2B}       \\
& ViT-L-14 &     \\
&  VIT-H-14 &    \\
\midrule 
 \multirow{4}{*}{UniCL}  
& \multirow{4}{*}{Swin-B}   & \tiny YFCC14M   & \multirow{4}{*}{ITC}  & \multirow{4}{*}{-}      \\ 
&  & \tiny IN21K           \\
&   & \tiny IN21K+YFCC14M \\
&   & \tiny IN21K+YFCC14M+GCC15M  \\
      \midrule  
 KLITE 
&              Swin-B                       & \tiny IN21K+YFCC14M+GCC15M & ITC & -   \\
\midrule 
   \multirow{2}{*}{BLIP}  
& \multirow{2}{*}{ViT-B-16}   & \multirow{2}{*}{\tiny\begin{tabular}[c]{@{}c@{}}  COCO+VG+CC+SBU\\+LAION+CapFilt-L \end{tabular}} & \multirow{2}{*}{\tiny\begin{tabular}[c]{@{}c@{}}ITC + ITM\\ + Captioning\end{tabular}} & -  \\
&  & &  & $\checkmark$  \\
 \midrule 
FLAVA  
& ViT-B/16       &          \tiny PMD70M     & \tiny\begin{tabular}[c]{@{}c@{}}ITC+ITM\\+MMM+MIM\\+MTM\end{tabular} & -         \\
\bottomrule
\end{tabular}
\end{scriptsize}
\end{center}
\end{table}

\section{Zero-shot Visual Recognition With Vision-Language Models}
In this study, we focus on two-stream contrastive vision-language models, such as CLIP, which leverage contrastive pre-training on a large dataset of paired image-text samples to learn cross-modal alignment. These models typically consist of a visual encoder $E_v$ and a text encoder $E_t$, for encoding visual inputs $x_v$ and textual inputs $x_t$ into aligned representation spaces.

The zero-shot visual recognition task with a vision-language model can be formulated as computing the cross-modality score:
\begin{equation}
\label{eq:score}
f(x_v, x_t) = E_v(x_v) \odot E_t(x_t)
\end{equation}
Here, the $\odot$ operator computes the score between visual and language embeddings, with cosine similarity being the common choice while some models like FLAVA use an additional module to fuse the multi-modal embeddings. For classification tasks, $x_t$ can be a class prompt, such as ``a photo of a {car}", or it can incorporate additional class-specific knowledge to improve performance. In our subsequent studies, we adopt the prompt templates used in ~\cite{radford2021clip} for classification tasks. We simplify $E_t(x_t)$ and $f(x_v, x_t)$ for a class $y$ to $E_t(y)$ and $f_{cls}(x_v, y)$, respectively.

In our study, we evaluate various contrastive vision-language models, each with distinct backbone architectures, training data, and learning objectives, shown in \cref{table:models}. These variants include CLIP~\cite{radford2021clip}, OpenCLIP~\cite{openclip} (trained on the public LAION dataset~\cite{schuhmann2022laionb}), UniCL~\cite{yang2022unicl} (which incorporates classification annotations into the contrastive learning objective), KLITE~\cite{shen2022klite} (which augments alt-text with extra knowledge during training), FLAVA~\cite{singh2022flava} (trained with both cross-modal and uni-modal data and losses), and BLIP~\cite{li2022blip} (which includes uni-modal and cross-modal training, along with a captioning head for data bootstrapping).
By examining these models, we aim to gain insights into the zero-shot visual recognition capabilities of vision-language models.

\section{Granularity Consistency of Vision-Language Models}
In this section, we investigate whether vision-language models (VLMs) perform consistently on visual concepts across different levels of granularity, which indicates their understanding of the relationships between concepts. We propose a benchmark to quantitatively evaluate the performance discrepancy of VLMs on concepts at different granularity levels. Our results show that models trained on image-text pairs exhibit significant performance discrepancies, with better recognition of moderately fine-grained concepts compared to coarse-grained ones. Further analysis suggests that the distribution of training data may account for this discrepancy, with models trained on datasets having more balanced representation across granularity levels showing smaller discrepancies. 

\subsection{Measure performance discrepancy on a semantic hierarchy}
\label{sec:multilabel_IN}

To assess the understanding of vision-language models across different levels of semantic granularity, we use zero-shot classification as our evaluation tool. Directly comparing classification metrics across granularities is not appropriate, as finer-grained classification inherently presents more challenges. Our benchmark focuses on measuring the discrepancy in zero-shot classification performance when using coarse-grained (CG) class prompts directly versus deriving predictions from using finer-grained (FG) children class prompts.

\paragraph{Dataset} 
We expand the popular ImageNet-1K dataset with multi-level label hierarchy. Each of the 1000 fine-grained labels is assigned its ancestor labels based on the WordNet hierarchy, adding 820 ancestor labels. For example, ``leopard" images are also labeled as ``big cat," ``feline," ``mammal," ``animal," and so on. his expansion allows us to thoroughly investigate how well VLMs perform across varying granularities.

\begin{figure}[t]
\centering
\begin{subfigure}{0.32\linewidth}
\includegraphics[width=\linewidth]{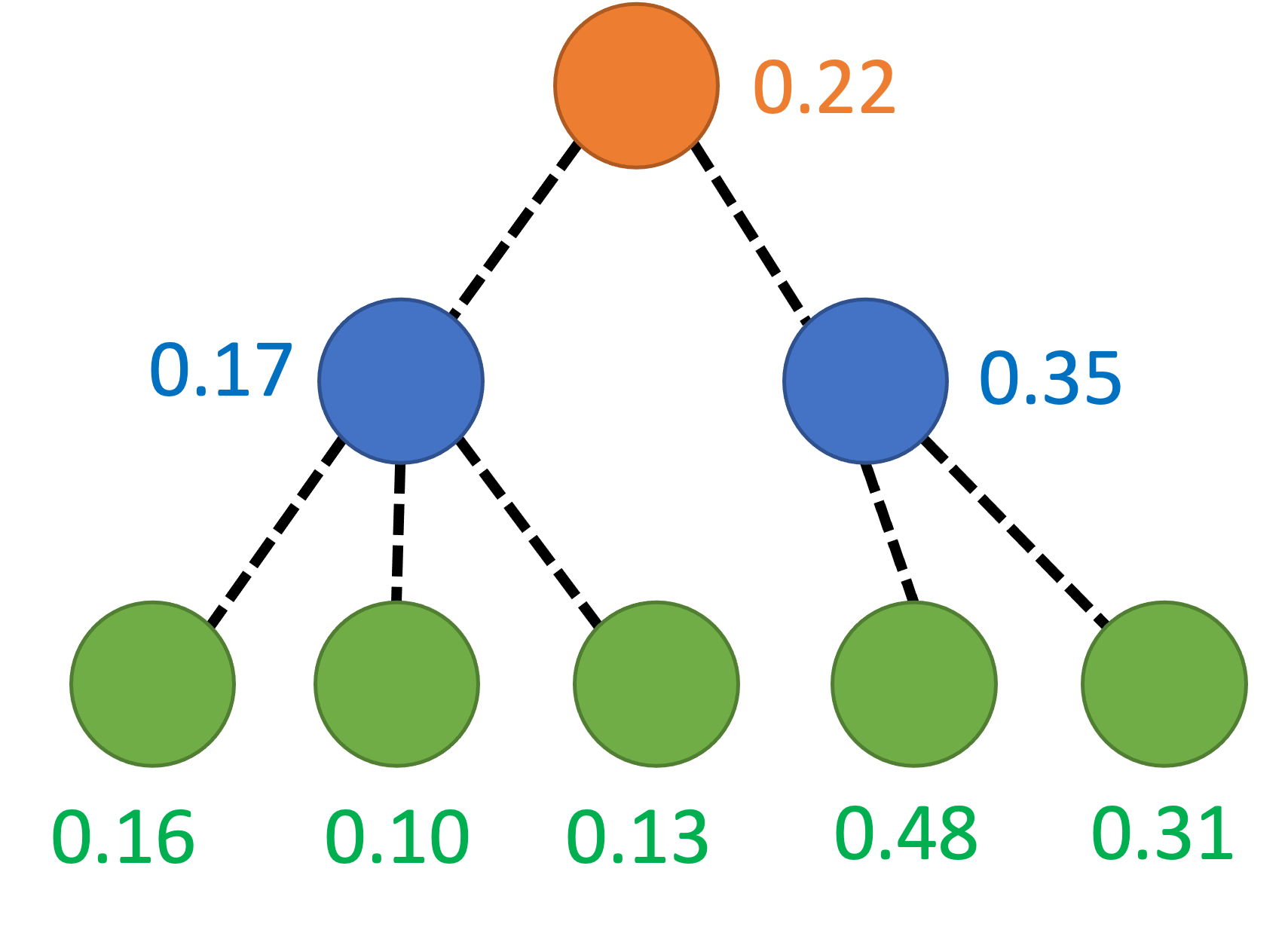}
\caption{\scriptsize Raw scores}
\label{fig:subim1}
\end{subfigure}
\begin{subfigure}{0.32\linewidth}
\includegraphics[width=\linewidth]{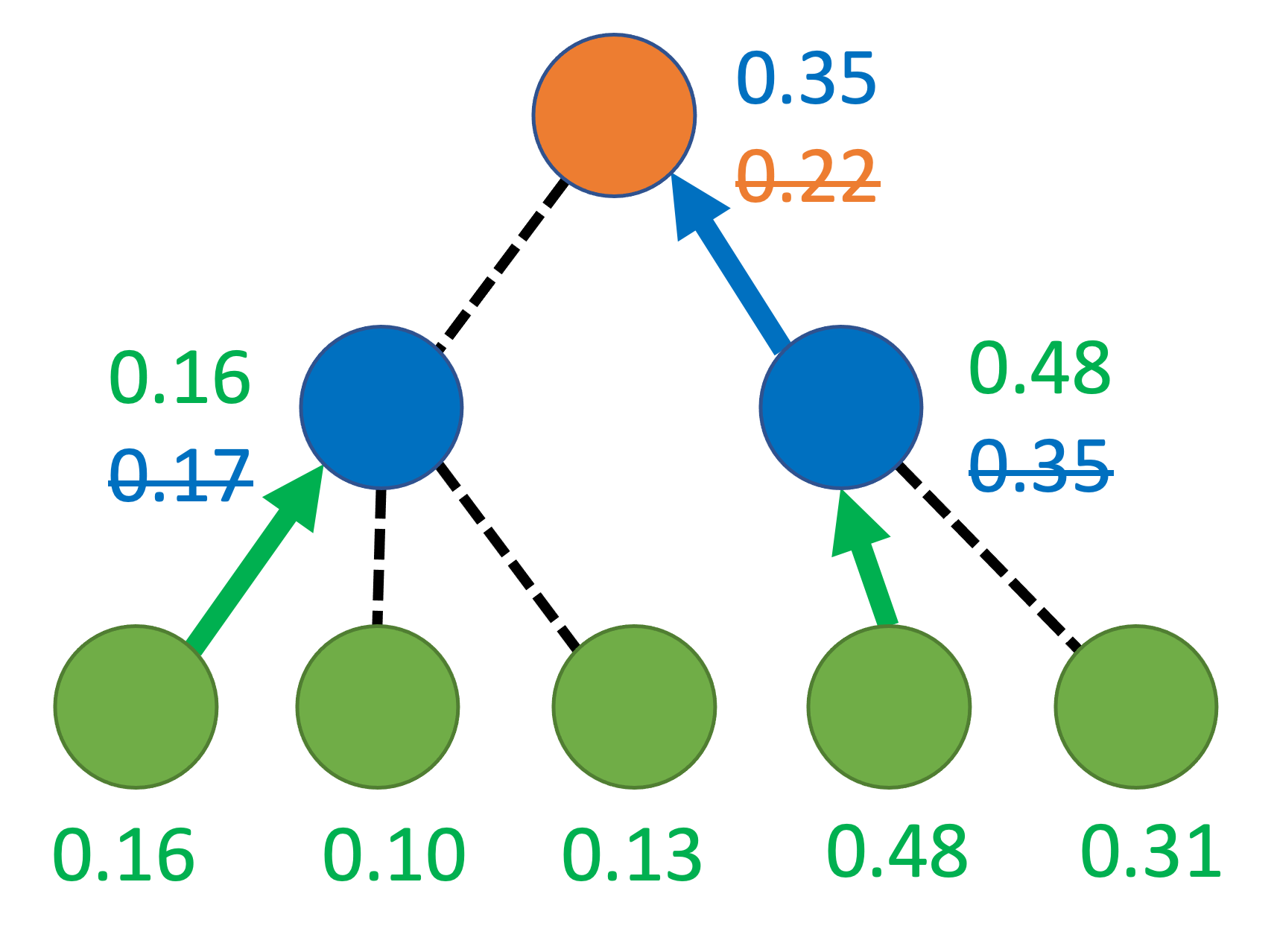}
\caption{\scriptsize Max of direct children}
\label{fig:subim2}
\end{subfigure}
\begin{subfigure}{0.3\linewidth}
\includegraphics[width=\linewidth]{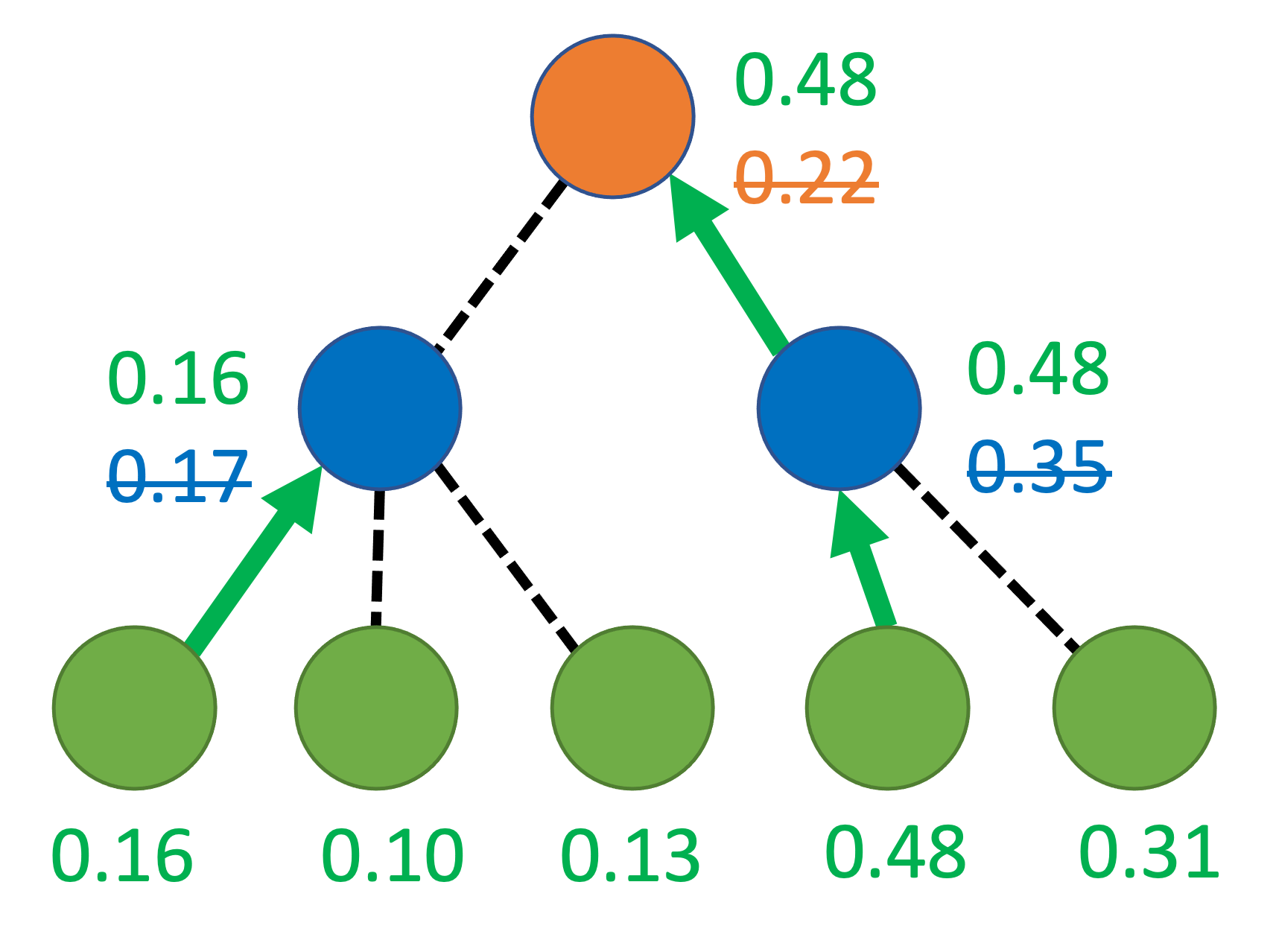}
\caption{\scriptsize Max of leaves}
\label{fig:subim3}
\end{subfigure}
\caption{Illustrations on the two ways to propagate scores on the semantic hierarchy. (a) Raw scores without propagation. (b) Propagate the max score from direct children classes. For example, 0.35 = max(0.17, 0.35) (c) Propagate the max score from leaf classes. For example, 0.48 = max(0.16, 0.10, 0.13, 0.48, 0.31)
}
\label{fig:multilevel_propagation}
\end{figure}

\paragraph{Evaluation Protocol} 
Given the hierarchical nature of our dataset, each image is associated with multiple labels, transforming our task into a multi-label classification setting. Here, each label is considered for binary classification independently. For ancestor (CG) labels, we employ two distinct methods for score prediction:
\begin{enumerate}[itemsep=0mm, parsep=0pt, leftmargin=*]
    \item Direct prediction: Utilize the text prompt of the CG label for cross-modality score with the image.
    \item 
Propagated prediction: Compute the aggregate scores from the FG children class prompts. For instance, the ``feline" label score can be derived by aggregating scores from ``lion", ``tiger", ``leopard", etc.
\end{enumerate}

For a class $y$, the raw cross-modality score between an image $x$ and the textual prompt of $y$ is computed by $S^{\text{raw}}(y) = f(x, y)$. For an ancestor class $y_i$, we design two specific approaches for score propagation, as illustrated in \cref{fig:multilevel_propagation} and formulated below:

\begin{enumerate}
    \item Propagate the maximum score from direct children classes.
    \begin{equation}
    S^{\text{child}}(y_i) = \max_{y_j\in Y_C^i}S^{\text{raw}}(y_j)   
    \end{equation}
    \item Propagate the maximum score from leaf (most fine-grained) chidren classes.
    \begin{equation}
  S^{\text{leaf}}(y_i) = \max_{y_j \in Y_C^i} S^{\text{leaf}}(y_j)
    \end{equation}
\end{enumerate}

The key idea is that if a VLM has a consistent understanding of concepts across granularities, its performance on directly classifying coarse-grained labels should be similar to the performance obtained by propagating predictions from the fine-grained children labels. We use mean Average Precision (mAP) as the evaluation metric, which does not require score threshold selection.  In \cref{table:multilabelIN}, we report mAP of leaf classes and ancestor classes. The performance of ancestor classes is evaluated with the raw scores, and scores using propagated from children classes using the two approaches presented above. The critical metric in our analysis is the difference ($\Delta$) between raw scores and propagated scores, serving as an indicator of how well the VLMs bridges the understanding between CG and FG concepts. 

\begin{table}
\caption{Zero-shot multi-label classification performance of labels at different levels of granularity on ImageNet. We reported the mean average precision (mAP) of ImageNet-1K fine-grained classes (leaves), and their coarse-grained ancestor classes with raw predictions ($\text{Anc}_{\text{raw}}$) and two propagated predictions. The differences ($\Delta$) between the raw and propagated performance of ancestor classes presents the performance discrepancy of vision-language models on concepts at different granularity. Propagating from leaf classes gives the best performance.}
\label{table:multilabelIN}
\begin{center}
\begin{footnotesize}
\begin{tabular}{lccccc}
\toprule
Config                         & Leaves & $\text{Anc}_{\text{raw}}$ & $\text{Anc}_{\text{child}}$ ($\Delta$) & $\text{Anc}_{\text{leaf}}$ ($\Delta$) \\ \midrule
\multicolumn{5}{c}{CLIP}                                                                                                                                            \\ \specialrule{0.1pt}{1pt}{1.5pt}
B-400M                         & 50.10  & 24.91                          & 45.35 \scriptsize(+20.44)                              & 58.73 \scriptsize(+33.83)                             \\
L-400M                         & 65.06  & 33.64                          & 57.72 \scriptsize(+24.08)                              & 72.25 \scriptsize(+38.61)                             \\ \midrule
\multicolumn{5}{c}{OpenCLIP}                                                                                                                                        \\ \specialrule{0.1pt}{1pt}{1.5pt}
B-400M                         & 47.10  & 20.12                          & 40.66 \scriptsize(+20.54)                              & 54.50 \scriptsize(+34.38)                             \\
B-2B                           & 54.97  & 24.95                          & 47.64 \scriptsize(+22.69)                              & 62.66 \scriptsize(+37.70)                             \\
L-2B                           & 65.79  & 31.59                          & 56.65 \scriptsize(+25.07)                              & 72.53 \scriptsize(+40.94)                             \\
H-2B                           & 68.28  & 32.70                          & 58.70 \scriptsize(+26.00)                              & 74.93 \scriptsize(+42.23)                             \\ \midrule
\multicolumn{5}{c}{UniCL(Swin-B)}                                                                                                                                           \\ \specialrule{0.1pt}{1pt}{1.5pt}
YFCC                   & 35.75  & 20.13                          & 35.90 \scriptsize(+15.77)                              & 47.55 \scriptsize(+27.42)                             \\
IN21K         & 26.28  & 38.15                          & 39.30 \scriptsize(+1.15)                               & 41.23 \scriptsize(+3.08)                              \\
\scriptsize{YFCC+IN21K}    & 37.84  & 35.18                          & 44.84 \scriptsize(+9.65)                               & 51.55 \scriptsize(+16.37)                             \\
All     & 54.49  & 37.54                          & 54.58 \scriptsize(+17.04)                              & 65.85 \scriptsize(+28.32)                             \\ \midrule
K-LITE                         & 48.40  & 31.50                          & 49.63 \scriptsize(+18.14)                              & 61.58 \scriptsize(+30.08)                             \\ \midrule
BLIP                           & 41.87  & 20.31                          & 39.44 \scriptsize(+19.13)                              & 52.08 \scriptsize(+31.77)                             \\
$\text{BLIP}_{\text{ft-coco}}$ & 42.83  & 22.07                          & 41.45 \scriptsize(+19.38)                              & 54.00 \scriptsize(+31.93)                             \\ \midrule
FLAVA                          & 40.91  & 21.36                          & 39.32 \scriptsize(+17.96)                              & 51.89 \scriptsize(+30.53)                             \\ 

\bottomrule
\end{tabular}
\end{footnotesize}
\end{center}
\end{table}

\begin{figure*}[ht]
\begin{minipage}{0.32\textwidth}
\includegraphics[width=\linewidth]{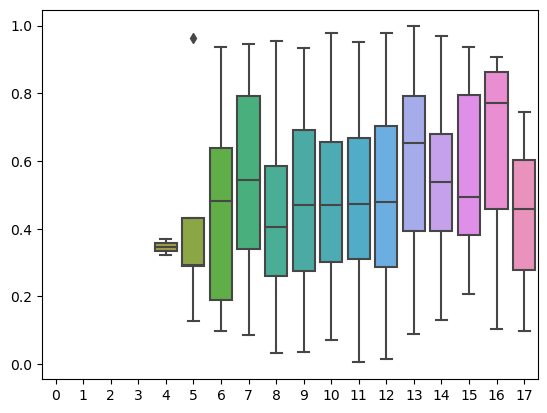}
\end{minipage}
\begin{minipage}{0.32\textwidth}
\includegraphics[width=\linewidth]{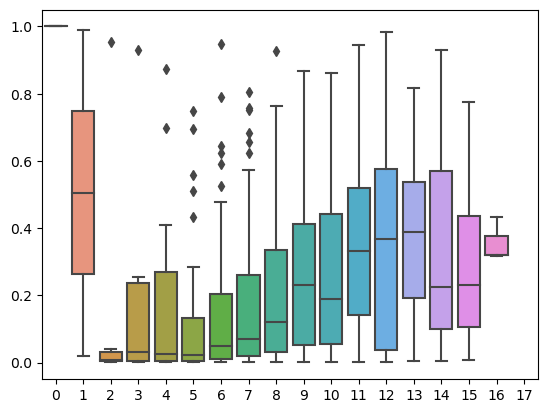}
\end{minipage}
\begin{minipage}{0.32\textwidth}
\includegraphics[width=\linewidth]{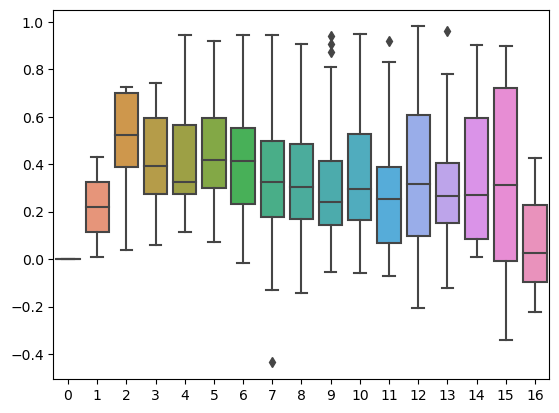}
\end{minipage}
\caption{\textbf{Left}: The box-plot of zero-shot classification performance (mAP) for leaf class over the level in the semantic hierarchy. \textbf{Middle}: The box-plot of classification performance (mAP) for ancestor classes over the level in the semantic tree. Note that level 0 and level 1 have 1 and 2 classes respectively and easy to get high mAP. \textbf{Right}:The box-plot of \textit{improved} zero-shot classification performance (mAP) for ancestor class by propagating from leaf classes, over the level in the semantic tree.}
\label{fig:multilabelIN_by_levels}
\end{figure*}

\subsection{Results and Analysis}
\label{sec:results_analysis}

\cref{table:multilabelIN} displays our granularity benchmark. The $\Delta$ values, highlighted in the last two columns, quantify the performance discrepancy between direct predictions using coarse-grained (CG) class prompts and predictions derived from prompts of their finer-grained (FG) children classes.

\textbf{General Trend Across Models:} Our analysis spans a diverse range of models, varying in scale and design. A significant trend emerges across all these models: direct predictions with CG labels consistently yield inferior results compared to those obtained from FG labels. Most notably, score propagation from the leaf (most fine-grained) classes leads to the most substantial performance improvements, decisively outperforming the propagation from direct children classes. This finding robustly indicates that VLMs demonstrate greater reliability and produce more accurate outputs when interacting with more specific, finer-grained concepts.

\textbf{Granularity-Level Performance Analysis:} 
Delving deeper into how VLMs perform at different granularity levels, we observe distinct patterns in their raw performance. These levels are defined based on the WordNet hierarchy, where level 0 represents the most abstract level ``entity" where all images are labeled with, and higher numbers indicate increasingly finer-grained concepts. For instance, at Level 3, a class might be as broad as ``signal" or ``location". At a middle level, such as Level 7, you might find classes like ``instrument" or ``vehicle." At even deeper levels, say Level 15, the classes are highly specific, like ``tiger shark" or ``cougar". According to \cref{fig:multilabelIN_by_levels}-Left and Middle, we find that VLMs are more adept at recognizing higher-level concepts for both leaf and ancestor classes. However, an intriguing dip in performance is noticed at the deepest level e.g. level 17 for leaf classes which consists of extremely fine-grained and likely rare concepts in the training data. These might be specific breeds of animals or types of vehicles not commonly encountered.
Interestingly, despite the challenges at the deepest level, propagating scores from leaf classes generally improves performance for the majority of CG ancestor classes. This improvement holds for 775 out of the 820 ancestor classes. The exception is noted at level 16, which does not benefit from propagation due to the underperformance of the level 17 child classes. The results in \cref{fig:multilabelIN_by_levels}-Right highlight these findings, illustrating that VLMs are best at recognizing \textit{moderatly} finegrained concepts

\begin{figure*}[t]
\begin{minipage}{0.45\textwidth}
\includegraphics[width=\linewidth]{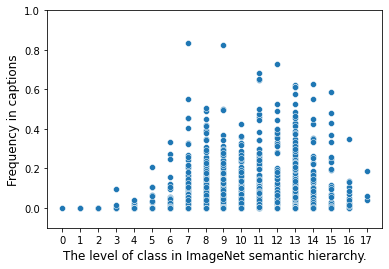}
\end{minipage}
\hfill  
\begin{minipage}{0.45\textwidth}
\includegraphics[width=\linewidth]{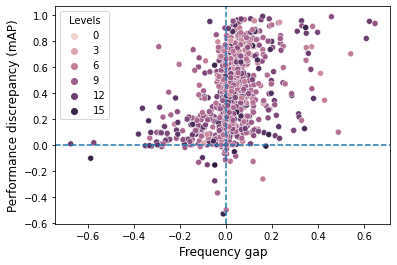}
\end{minipage}
\caption{\textbf{Left}: The scatter-plot of the frequency of class names in pre-training captions over the level in the semantic tree. Course-grained and overly fine-grained concepts are less presented in captions. \textbf{Right}: The scatter-plot of performance discrepancy over the frequency gap between ancestor class names and their leaf children. A positive correlation exists between the performance discrepancy and frequency gap (coefficient 0.43 with p-value 3.4e-39). }
\label{fig:freq_performance correlation}
\end{figure*}
 
\textbf{Influence of Pre-training Data on Granularity Discrepancy:}
Contrary to expectations, scaling up the alt-text training data or increasing model sizes, such as in the case of OpenCLIP-H-2B, does not seem to effectively address the granularity discrepancy. In fact, such scaling might worsen it. \textit{The distribution of training data content is a more critical factor than its volume in influencing the granularity discrepancy}. This becomes evident when comparing the performance of UniCL models trained on different datasets, such as ImageNet21K, YFCC-14M, and GCC-15M. 
\begin{enumerate}[itemsep=0mm, parsep=0pt, leftmargin=*]
\item UniCL models trained on image alt-text data (UniCL${\text{YFCC}}$) significantly outperform those trained on ImageNet21K (UniCL${\text{IN21K}}$) in FG leaf label classification (35.75 vs. 26.28 mAP).
\item UniCL$_{\text{IN21K}}$ excels in CG ancestor raw performance due to the comprehensive inclusion of CG classes in IN21K, even surpassing CLIP and OpenCLIP models trained on larger-scale alt-text datasets.
\item However, integrating alt-text data (YFCC14M and GCC15M) with IN21K for UniCL training enhances FG classification at the cost of CG performance, leading to a larger CG-FG discrepancy.
\end{enumerate}

\textbf{Granularity Bias in Pre-training Data:}  The distribution of alt-text data, skewed towards fine-grained concepts, appears to contribute significantly to the observed performance discrepancy. The natural inclination to use precise concepts in language descriptions appears to be a driving factor behind this bias. In our further analysis of the OpenCLIP models trained on LAION-2B, we investigate how the distribution of visual concepts in alt-text data correlates with the granularity discrepancy. We use ImageNet samples from each leaf class to find similar images in LAION-2B, determining the frequency of each class name in the training captions. This frequency distribution in \cref{fig:freq_performance correlation}-Left shows that higher-level (more fine-grained) classes are mentioned more frequently except for the overly fine-grained classes (level$\geq$16), aligning with the performance trends observed in \cref{fig:multilabelIN_by_levels}. Moreover, we examine the correlation between each ancestor class's performance discrepancy ($\Delta_{\text{leaf}}$) and its frequency gap with its leaf children ($\Delta_{\text{freq}}$). Our findings, shown in \cref{fig:freq_performance correlation}-Right, demonstrate a positive correlation (coefficient 0.43 with a significant p-value of $3.4e-39$), reinforcing the notion that the distribution of training data significantly influences the VLMs' granularity bias.

In conclusion, the insights from our granularity benchmark shed light on the significant challenges and limitations inherent in current VLMs, particularly their inconsistent performance across different levels of semantic granularity. These results highlight the imperative for more balanced and diverse training datasets in developing VLMs capable of robust

\section{Evaluate Specificity Robustness}

\begin{figure*}[ht]
\begin{minipage}{0.32\textwidth}
\label{fig:coco_pos}
\includegraphics[width=\linewidth]{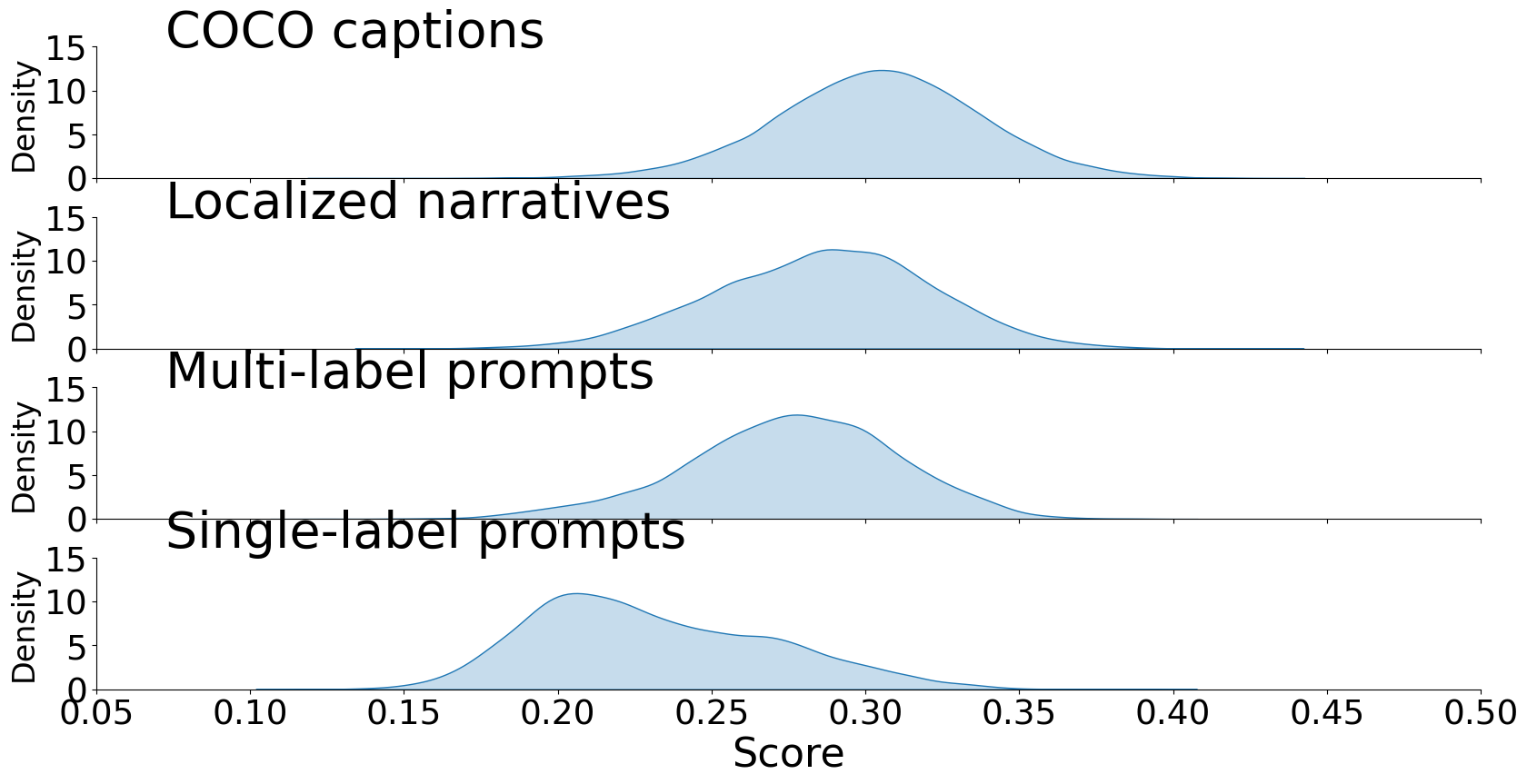}
\end{minipage}
\begin{minipage}{0.32\textwidth}
\includegraphics[width=\linewidth]{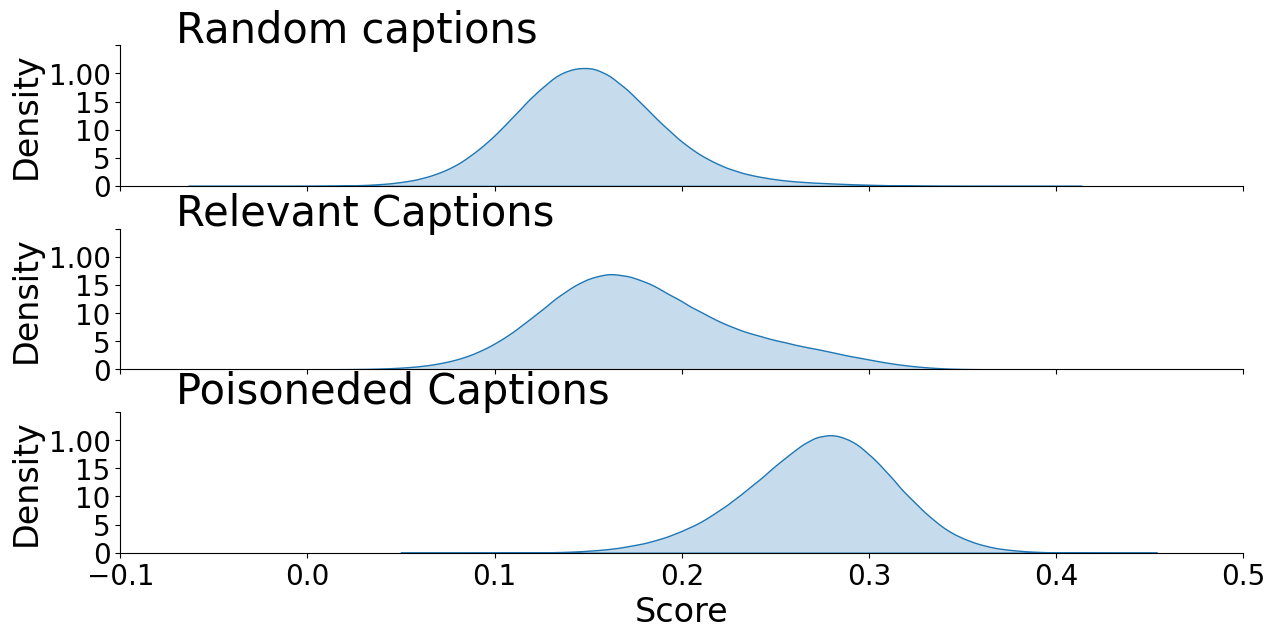}
\end{minipage}
\begin{minipage}{0.32\textwidth}
\includegraphics[width=\linewidth]{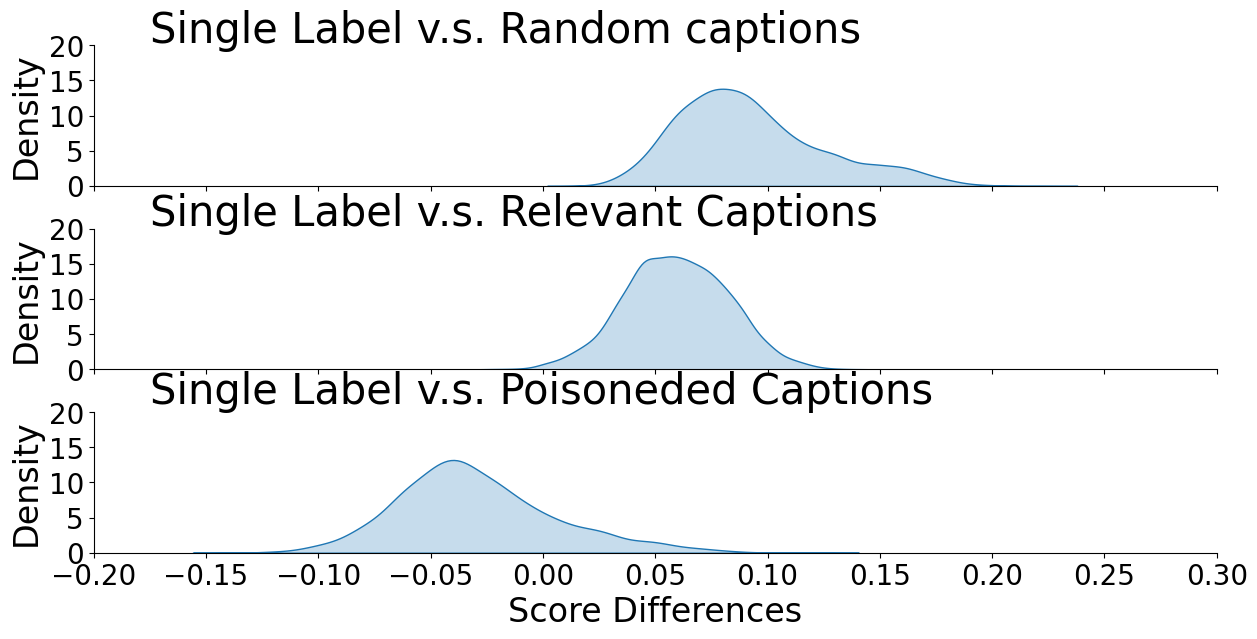}
\end{minipage}
\caption{\textbf{Left}: Distribution of cross-modality scores with positive text: COCO captions, Localized-narratives captions, single-label, and multi-label prompts. Mismatched specificity in text (either too low or high) results in reduced scores. \textbf{Middle}: Distribution of scores with negative text: captions from random images, relevant images, and subtly altered captions. The altered captions attain high scores, similar to positive texts. \textbf{Right}: Score differences between single-label prompts and various negative texts, highlighting that correct single-label prompts often score lower than incorrect altered captions.}  \label{fig:coco_scores}
\end{figure*}

\begin{table*}[htp]
\caption{Performance of image-to-text retrieval measured by mean Average Precision (mAP). Columns differentiate between positive and negative text types: $\text{Cap}^+$ (true COCO captions), $\text{Cap}^+_{ln}$ (localized-narratives captions), $\text{Prompt}^+_s$ and $\text{Prompt}^+_m$ (single/multi-label prompts) for positives; $\text{Cap}^-_{rd}$, $\text{Cap}^-_{rl}$, $\text{Cap}^-_{er}$ (captions from random images, relevant images, and error-modified true captions) for negatives. This shows how hard positives and negatives can distort vision-language models' similarity scores.}
\label{table:coco_img2txt_retrieval}
\begin{center}
\begin{footnotesize}
\begin{tabular}{llll|lll|lll|lll}
\toprule
\multirow{2}{*}{Model} &   \multicolumn{3}{c}{$\text{Cap}^+$} &
   \multicolumn{3}{c}{$\text{Prompt}^+_m$}  & \multicolumn{3}{c}{$\text{Prompt}^+_s$} & \multicolumn{3}{c}{$\text{Cap}^+_{ln}$} \\
\cmidrule{2-13} 

 & $\text{Cap}^-_{rd}$ & $\text{Cap}^-_{rl}$ & $\text{Cap}^-_{er}$ 
 & $\text{Cap}^-_{rd}$ & $\text{Cap}^-_{rl}$ & $\text{Cap}^-_{er}$ 
 & $\text{Cap}^-_{rd}$ & $\text{Cap}^-_{rl}$ & $\text{Cap}^-_{er}$ 
 & $\text{Cap}^-_{rd}$ & $\text{Cap}^-_{rl}$ & $\text{Cap}^-_{er}$ \\

\midrule
CLIP-B             & 94.78          & 82.77          & 28.10           & 74.39          & 50.12          & 7.19           & 55.69          & 30.17          & 4.47 
& 81.29 &	60.00 &	13.57\\
CLIP-L             & 95.64          & 84.66          & 30.59          & 79.84          & 56.76          & 8.51           & 58.18          & 33.52          & 4.94           & 85.51 &	66.74 &	16.63  \\
\midrule
OpenCLIP$_{\text{B-400M}}$  & 95.28                & 84.62                & 29.61                     & 64.66                    & 39.1                      & 4.49                     & 50.9                      & 25.93                     & 3.63                     & 83.48                     & 63.07                     & 13.85               
\\
OpenCLIP$_{\text{B-2B}}$  & 96.28                & 86.73                & 28.96                     & 75.84                    & 51.83                     & 6.32                     & 61.39                     & 35.89                     & 4.42                     & 88.83                     & 71.76                     & 18.91           
 \\
OpenCLIP$_{\text{L-2B}}$ & 97.09          & 88.81          & 33.03          & 79.22          & 56.00             & 6.90            & 65.44          & 39.97          & 4.96   &  89.50&	72.78&18.63      \\
OpenCLIP$_{\text{H-2B}}$ & 97.45                & 89.85                & 35.82                     & 79.2                     & 57.64                     & 7.49                     & 65.67                     & 42.19                     & 5.75                     & 89.74                     & 73.28                     & 18.09               
 \\
\midrule

UniCL$_{\text{All}}$                  & 94.37          & 81.76          & 20.74          & 82.58          & 62.33          & 9.94           & \textbf{82.45} & \textbf{60.02} & 8.71 &    81.96&	62.33&	12.99      \\
\midrule

KLITE                     & 92.47          & 77.67          & 16.45          & 75.71          & 53.60           & 9.03           & 69.98          & 47.06          & 8.47    &  79.81&	59.24&	11.16     \\
\midrule

BLIP                      & 97.68          & 90.89          & 48.53          & 57.64          & 32.21          & 3.13           & 43.24          & 20.07          & 2.81     & 82.26&	63.62	&17.94     \\
BLIP$_{\text{ft-coco}}$              & {99.07} & 95.15 & \textbf{56.44} & 74.65          & 51.02          & 4.86           & 65.96          & 41.77          & 4.02    &  89.99&	75.92&	23.13     \\
BLIP$_{\text{ft-coco-fusion}}$              & \textbf{99.26}                & \textbf{96.08}                & 38.57                     & 76.59                    & 54.97                     & 3.35                     & 81.62                     & 58.41                     & 2.97                     & 92.51                     & 82.59                     & 22.72                   \\
\midrule

FLAVA                     & 97.73          & 89.31          & 29.49          & \textbf{86.52} & \textbf{69.29} & \textbf{13.22} & 78.35          & 58.33          & \textbf{11.22} & \textbf{94.83} &	\textbf{82.87} &	\textbf{35.09} \\
\midrule
NegCLIP  & 96.6	& 87.37	& 51.88	& 65.32	& 39.91	& 6.7 & 	61.32& 	34.52& 	6.09 & 76.70&	53.93&	13.33 \\
\bottomrule
\end{tabular}
\end{footnotesize}
\end{center}
\end{table*}

When using vision-language models for open-world zero-shot recognition, the textual describes the visual concepts to recognize and the output score should indicate the chance that the described concepts exist in the visual input. In other words, it is critical to measure the correctness of textual inputs given visual inputs. However, as the example in \cref{fig:paper_concept}-Right illustrates, the scores of visual language models and do not strictly reflect the correctness of the textual input and thus make it challenging to be useful for open-world visual recognition.  Since contrastive vision-language models have been trained on image alt-text pairs, the scores are biased toward the specificity of text as in the pretraining data. In our study, we demonstrated that the specificity of text can distract vision-language scores that VLMs struggle to reflect the correctness faithfully.

\paragraph{Evaluation protocol and dataset} We use image-to-text retrieval as the proxy task to demonstrate that the scores of contrastive vision language models can easily be distracted. We build our experiments on images of the MSCOCO2017 dataset and their annotation of captions and bounding boxes. The setup of the image-to-text retrieval task is following. Given a query image and a set of positive and negative text, the score between the query image and each text is used for retrieving the positive text. Average Precision (AP) is the metric for evaluating the performance of each image and we report the mean Average Precision (mAP) of the whole dataset. Typically, the positive text are the captions annotated for the query images ($\text{Cap}^+$), and the negative text is the captions of other images in the data ($\text{Cap}^-_{rd}$). To test our hypothesis, we design the following hard positives and hard negatives.  
\begin{itemize}
    \item \textit{Prompts of a single label} ($\text{Prompt}^+_s$): apply the classification prompts on one label of the query image. For example, ``a photo of a dog". 
    \item \textit{Prompts of multiple labels} ($\text{Prompt}^+_m$): apply the classification prompts on all labels in the query image. For example, ``a photo of dog, person, ball".
    \item \textit{Captions from Localized narratives\cite{PontTuset_eccv2020}} ($\text{Cap}^+_ln$): the text descriptions that are much longer and more informative than typical captions in MSCOCO and pretraining data.
    \item \textit{Captions of relevant images} ($\text{Cap}^-_{rl}$): COCO captions of relevant images that have overlapping labels with the query image.
    \item \textit{Captions with errors} ($\text{Cap}^-_{er}$): modifying true COCO captions of query images with errors by replacing a noun entity in the text with the name of a label that does not exit in the image. We use spaCy \footnote{\url{https://spacy.io/}} for entity recognition.
\end{itemize}

The hard positives $\text{Prompt}^+_s$, $\text{Prompt}^+_m$ and $\text{Cap}^+_ln$ contain less or more information, although still correct, than the true captions $\text{Cap}^+$. They can examine if different specificity of the positive text can reduce the score. The hard negatives $\text{Cap}^-_{rl}$ and $\text{Cap}^-_{er}$ are similar to true captions but are wrong descriptions. They can examine whether the model can be robust to specificity and indicate the correctness of text input. Note that we use randomly chosen 100 negative texts for each query image for all image-text retrieval experiments and report results of CLIP ViT-B/32.

\begin{table*}[t]
\caption{Performance of fine-tuned CLIP-B models in MSCOCO-2017's image-to-text retrieval, measured by mean Average Precision.}
\vskip -0.1in
\label{table:finetuning}
\begin{center}
\begin{footnotesize}
\begin{tabular}{llll|lll|lll|lll}
\toprule
\multirow{2}{*}{Model} &   \multicolumn{3}{c}{$\text{Cap}^+$} &
   \multicolumn{3}{c}{$\text{Prompt}^+_m$}  & \multicolumn{3}{c}{$\text{Prompt}^+_s$} & \multicolumn{3}{c}{$\text{Cap}^+_{ln}$} \\
\cmidrule{2-13} 

 & $\text{Cap}^-_{rd}$ & $\text{Cap}^-_{rl}$ & $\text{Cap}^-_{er}$ 
 & $\text{Cap}^-_{rd}$ & $\text{Cap}^-_{rl}$ & $\text{Cap}^-_{er}$ 
 & $\text{Cap}^-_{rd}$ & $\text{Cap}^-_{rl}$ & $\text{Cap}^-_{er}$ 
 & $\text{Cap}^-_{rd}$ & $\text{Cap}^-_{rl}$ & $\text{Cap}^-_{er}$ \\
 
\midrule
CLIP-B           & 94.78          & 82.77          & 28.1           & 74.39          & 50.12          & 7.19           & 55.69          & 30.17          & 4.47    & \textbf{81.29}  &	\textbf{60.00} &	13.57  \\
CLIP-B$_{\text{ft-coco}}$         & \textbf{96.98} & \textbf{88.15} & 35.04          & 71.65          & 47.16          & 5.41           & 60.4           & 33.53          & 4.09   & 80.52 &	57.96 &	10.66  \\
NegCLIP         & 96.6           & 87.37          & 51.88          & 65.32          & 39.91          & 6.7            & 61.32          & 34.52          & 6.09    & 76.70&	53.93&	13.33  \\
Ours & 96.21          & 85.9           & \textbf{75.74} & \textbf{93.37} & \textbf{75.46} & \textbf{55.95} & \textbf{83.22} & \textbf{57.98} & \textbf{29.05}  & 78.54 &	55.11 &	\textbf{31.74}\\

\bottomrule
\end{tabular}
\end{footnotesize}
\end{center}
\vskip -0.1in
\end{table*}

\paragraph{Results and implications} We first plot a normalized histogram of visual-language scores between query images and various textual inputs in \cref{fig:coco_scores}. \cref{fig:coco_scores}-Left compares the scores from different types of positive texts. True COCO captions generate higher score than classification prompts (multiple-label prompts get higher scores than single-label prompts), and Localized-narratives who are overly-detailed captions surprisingly lead to lower scores than normal captions. The observations confirms our hypothesis that \textit{the amount of information (specificity) in text can distort the scores} and the specificity that is closer to the training text leads to higher scores. Shown by \cref{fig:coco_scores}-Middle, relevant image captions score slightly higher than random ones among negative texts. However, captions with modified errors score as high as true captions, undermining the effectiveness of VL scores. The result verifies our hypothesis that \textit{the similarity scores cannot distinguish the correctness}. When comparing the positive single-label prompts with different types of negative text in \cref{fig:coco_scores}-Right, single label positives $\text{Prompt}^+_s$ are even lower than hard negatives $\text{Cap}^-_{er}$, which is not desired.

Then, we report the image-to-text retrieval results in \cref{table:coco_img2txt_retrieval} when combining different positive and negative text. We can see that using harder positives or harder negatives can degrade image-to-text retrieval performance, and retrieving label prompts from captions with small errors is extremely hard. Comparing the performance of different models, we can see that the BLIP model with the fusion design fine-tuned on COCO is the best when the positive text are true captions which is nature since it is trained on the same data. however, results in worse performance when distinguishing poisoned captions.  When the positive text is label prompts, FLAVA is the best or the second best model, probabally due to its additional uni-modal training data/loss. UniCL is the best when single-label prompts are the positives, which we think can be explained by the ImageNet21K classification dataset in its training data.

\section{Limitations of Fine-Tuning VLMs with Hard Samples}
In our study, we generate both hard positive and negative text samples to examine whether vision-and-language models (VLMs) accurately interpret text that aptly describes images. In this section, we explore the efficacy of using these hard text samples for training or fine-tuning VLMs. Drawing inspiration from NegCLIP in a recent study \cite{yuksekgonul2022vlmbow}, which delves into understanding compositional relationships in VLMs, we fine-tune the CLIP-B model using these challenging samples on MSCOCO training data. Our approach differs from NegCLIP in that we focus exclusively on hard text samples for fine-tuning. These include complex positives such as single/multi-label prompts and difficult negatives like captions from relevant images or true captions subtly modified with errors, forming part of our benchmarking strategy. Notably, NegCLIP also uses modified true captions with errors to test compositional understanding.

Following the default fine-tuning hyperparameters cited in \cite{yuksekgonul2022vlmbow}, we additionally fine-tune CLIP-B with original MSCOCO captions (CLIP-B${\text{ft-coco}}$) to establish a fairer baseline. The performance of our models, alongside NegCLIP, is detailed in Table \ref{table:finetuning}. Our findings indicate that fine-tuning with original MSCOCO data without hard samples shows superior performance in distinguishing true captions from simpler negative ones, such as $\text{Cap}^+$ vs $\text{Cap}^-_{rd}$ or $\text{Cap}^-_{rl}$. NegCLIP surpasses the COCO-fine-tuned CLIP only in tasks aligned with its data augmentation strategy, e.g., $\text{Cap}^+$ vs $\text{Cap}^-_{er}$, but shows similar or even reduced effectiveness in other tasks. Importantly, our fine-tuned model demonstrates superior performance compared to the baselines and NegCLIP in the most challenging scenarios, $\text{Prompt}^+_s$ vs $\text{Cap}^-_{er}$, owing to a broader coverage of harder text sample types.

Despite these enhancements, challenges remain. NegCLIP and our fine-tuned model struggle in complex scenarios like single/multi-label prompts ($\text{Prompt}^+_{s}$ and $\text{Prompt}^+_{m}$) against subtly altered captions $\text{Cap}^-_{er}$. For tasks with unseen hard text types, like long captions $\text{Cap}^+_{ln}$ vs $\text{Cap}^-_{rd}$ and $\text{Cap}^-_{rl}$, all fine-tuned models underperform compared to the original CLIP. This indicates that while fine-tuning with hard samples enhances performance on familiar challenging cases, it falls short in addressing the full spectrum of difficulties, especially when encountering cases outside our augmentation strategy. These findings underscore the limitations of solely relying on fine-tuning with hard samples and highlight the urgency for a more comprehensive solution capable of encompassing a wider variety of potential scenarios, as our current methods do not completely resolve the issue.

\section{Conclusion and Discussion}
As interest in vision-language models grows, we present a novel benchmark and comprehensive study on the behaviors that create challenges to be useful in the open-world settings. First, we demonstrate that VLMs perform inconsistently on concepts of different semantic granularity. Based on our experiments, the performance discrepancy is due to the biased distribution of training data. Second, we show that vision language scores mostly measure similarity rather than correctness and can be distracted by the specificity of text. The scores are higher when the specificity of text are closer to the captions in the training data. This issue cannot be systematically solved by fine-tuning wit hard text mining.

While our study doesn't offer complete solutions for the identified issues, we suggest several promising avenues for improvement. Firstly, addressing the granularity discrepancy and specificity sensitivity could involve enhancing the training data. This can be achieved by augmenting text with a more balanced concept distribution and incorporating hard negatives and positives, possibly using large language models for assistance. Secondly, the dual encoder and embedding similarity approach inherently complicates correct recognition. For instance, true captions and slightly erroneous ones may have similar embeddings from a uni-modal view, resulting in close scores with identical image embeddings. A more advanced cross-modality fusion module could be key in discerning between visual and textual features, enabling distinct outputs for similar textual inputs. Lastly, large language models (LLMs) trained on more diverse text data might help mitigate the challenges we've noted. Our language-only experiment in the Appendix illustrates the potential of using generative LLMs in this context. Exploring and evaluating VLMs integrated with generative LLMs, e.g. vision-LLM, for recognition tasks represents an exciting future direction.

{
    \small
    \bibliographystyle{ieeenat_fullname}
    \bibliography{vlm}
}


\clearpage
\newpage
\appendix

\input{supplementary}

\end{document}

%% file: preamble.tex
\usepackage[dvipsnames]{xcolor}


%% file: supplementary.tex
\section{A Two-level Granularity Benchmark}
In this section, we presents an simplified granularity benchmark with two-levels of semantic hierarchy. The results are consistent with our observations in the main paper.

\paragraph{Two-level Dataset} Our evaluation starts on a dataset with two levels of labels: $N_{cg}$ coarse-grained (CG) classes $Y_{cg} = \{y_{cg}^i\}$, where $i\in\{1,...,N_{cg}\}$, and each CG class has $N^i_{fg}$ fine-grained (FG) children classes $Y_{fg}^i = \{y_{fg}^{i,j}\}$, where $j\in\{1,...,N^i_{fg}\}$. In total, there are $N_{fg} = \sum_{i=1}^{N_{cg}}N_{fg}^i$ FG classes. To create our two-level classification dataset, we adapt the tiered-ImageNet \cite{ren2018meta} benchmark, which has 608 FG classes (a subset of the original 1000 classes of ImageNet-1K) organized under 34 CG classes and covers 30,400 out of 50,000 ILSVRC-12 validation images.

\paragraph{Evlauation protocol} For \textit{two-level} granularity, we measure the performance difference of CG classification between using direct predictions with CG prompts and propagated FG predictions. The simplest propagation method is to assign the predicted FG labels to their CG parents' labels. For instance, if an image is predicted as "golden retriever" in the FG classification, it is labeled with its CG parent class "animal." Intuitively, if a model exhibits consistent understanding of CG and FG concepts, the performance of CG classification using CG prompts should be similar to propagating the results from FG classification. An alternative way of propagating FG to CG concepts is using the aggregated embeddings of FG prompts for CG classifcation. Specifically, for the $i$-th CG class, we compute the average of the FG prompt embeddings as the CG prompt embedding: $E^{prop}_t(y^i_{cg}) = \frac{1}{N^i_{fg}}\sum_{j=1}^{N_{fg}^i}E_t(y_{fg}^{i,j})
$. We use top1 accuracy as the classification metric.

\begin{table*}[htp]
\caption{Evaluating vision-Language model zero-shot classification performance (top-1 accuracy) on fine-grained classes (FG) and coarse-grained (CG) classes. The CG classification results are obtained through two methods: relating predicted FG class labels to their CG parents (CG$_{\text{FG-label}}$) and using the average of the FG prompt embeddings as the CG prompt embedding (CG$_{\text{FG-emb}}$). We measure the differences ($\Delta$) with CG classification using CG class prompts (CG$_{\text{direct}}$), which reveals the discrepancy in CG-FG performance of vision-language models.}
\label{table:FG2CG}
\vskip 0.15in
\begin{center}
\begin{tabular}{lllllll}
\toprule
Model & Arch & Training data & FG$_{\text{direct}}$ & CG$_{\text{direct}}$ & CG$_{\text{FG-label}}$ ($\Delta$) & CG$_{\text{FG-emb}}$ ($\Delta$) \\
\midrule 
CLIP  & \tiny ViT-B-32 & \tiny Private400M      & 66.47 & 50.15 & 86.35 (+36.2)  & 72.62 (+22.47)  \\
\midrule
Open-CLIP & \tiny ViT-B-32 & \tiny LAION400M & 63.82 & 35.98 & 84.08 (+48.1)  & 69.65 (+33.67) \\
          & \tiny ViT-B-32 & \tiny LAION2B   & 69.78 & 45.54 & 87.39 (+41.85) & 71.54 (+26)       \\
          & \tiny ViT-L-14 & \tiny LAION2B   & 77.72 & 49.74 & 91.83 (+42.09) & 76.49 (+26.75)  \\
          & \tiny VIT-H-14 & \tiny LAION2B   & 80.39 & 52.22 & 92.86 (+40.64) & 77.43 (+25.21)  \\
\midrule 
UniCL & \tiny Swin-B & \tiny YFCC14M           & 41.14 & 37.37 & 69.67 (+32.3)  & 59.75 (+22.38)  \\ 
      & \tiny Swin-B & \tiny IN21K              & 30.6  & 53.14 & 66.26 (+13.12) & 59.5 (+6.36)  \\
      & \tiny Swin-B & \tiny IN21K+YFCC14M   & 45.91 & 52.27 & 76.84 (+24.57)   & 67.63 (+15.36)  \\
      & \tiny Swin-B & {\tiny IN21K+YFCC14M+GCC15M} & 60.17 & 51.9  & 83.44 (+31.54) & 68.37 (+16.47)  \\
      \midrule 

K-LITE                & \tiny Swin-B                       & \tiny IN21K+YFCC14M+GCC15M & 54.75 & 44.92 & 81.85 (+36.93) & 71.05 (+26.13) \\
\midrule 
BLIP &  \multirow{2}{*}{\tiny ViT-B-16}   & \multirow{2}{*}{\shortstack[l]{\tiny COCO+VG+CC+SBU\\\tiny +LAION+CapFilt-L}}   & 55.41 & 42.09 & 80.92 (+38.83) & 69.69 (+27.6)  \\
$\text{BLIP}_{\text{ft-coco}}$                      &         &                         & 58.02 & 46.75 & 84.7 (+37.95) & 72.93 (+26.18) \\
 \midrule 
FLAVA                 & \tiny ViT-B/16       &          \tiny PMD70M               & 59.48 & 50.11 & 83.37(+33.26) & 70.84 (+20.73) \\
\bottomrule
\end{tabular}
\end{center}
\vskip -0.1in
\end{table*}

\section{A Language Only Study}
\label{sec:language}
In the main paper, we have highlighted the issues faced by vision and language models (VLMs) in zero-shot recognition tasks, focusing on both granularity and correctness analyses. Since these analyses primarily involve working with different text inputs while keeping the visual inputs constant, improving the language encoder becomes a natural next step. We address the question of whether language embeddings from pre-trained large-scale language models (LLMs) exhibit better behavior compared to VLMs. To investigate this, we design a language-only task.

Specifically, we conduct a text classification task that involves classifying fine-grained (FG) concepts to their corresponding coarse-grained (CG) concepts using the same two-level ImageNet dataset as in Section 4.1. This results in a 34-way classification task with 608 text samples (FG concept prompts). Similar to zero-shot image classification, we compute the cosine similarity between the language embeddings of FG and CG prompts and classify a FG concept to the CG concept with the highest similarity score. To incorporate the generative model GPT-3 for this task, we design the following zero-shot prompt:

\begin{quotation}
\noindent "Classify a given concept into one of the following classes: \$\{\textit{all coarse-grained concepts} \}. \\
Q: \$\{\textit{a fine-grained\_concept}\} A:" 
\end{quotation}

\cref{table:txt2txt} Presents the performance of LLMs\footnote{We use pretrained models provided by \texttt{sentence-transformer} \url{https://github.com/UKPLab/sentence-transformers}} or the language encoder of VLMs on the language-only task. Surprisingly, LLMs, even when fine-tuned for sentence embedding, do not outperform the language encoder of VLMs. However, we find that GPT-3 performs significantly better in a generative manner. This suggests that when dealing with concept relationships on a larger scale where simple embedding similarity struggles, generative modeling may offer a more powerful approach to capture complex semantic knowledge and model the relationships effectively. 

\begin{table*}[htp]
\caption{Performance (accuracy) of classify a fine-grained concept to coarse-grain concept using language embedding models or generative language models.}
\label{table:txt2txt}
\begin{center}
\begin{small}
\begin{tabular}{lc}
\toprule
Model Type   & FG-to-CG Text Classification Accuracy (\%) \\
\midrule
CLIP-B   &   61.18             \\
OpenCLIP-L$_{\text{LAION2B}}$  &  55.76      \\
OpenCLIP-H$_{\text{LAION2B}}$  &  62.66   \\
UniCL   &      52.96                   \\
KLITE   &      43.59                   \\
BLIP    &      50.00                      \\
FLAVA    &      57.40      \\
\midrule
all-roberta-large-v1 &  51.81 \\
sentence-T5-large   &   52.47 \\
sentence-T5-xl   &   55.26\\
\midrule
GPT-3$_{\text{text-davinci-002}}$   & 71.17 \\
\bottomrule
\end{tabular}
\end{small}
\end{center}
\end{table*}

\section{Limitations of Our Study}

While our study provides valuable insights into the challenges and limitations of vision-and-language models (VLMs) for zero-shot visual recognition, it is important to acknowledge several limitations. Firstly, our experiments primarily focus on a specific set of VLMs, datasets, and evaluation metrics. While we have made efforts to select representative models and datasets, our findings may not fully generalize to the entire landscape of vision and language models. Generalizing the results to other VLM architectures or datasets requires further investigation and experimentation.

Secondly, our study is conducted within the context of the evaluation protocols and benchmarks we have proposed. While we have designed these protocols to address the challenges of zero-shot recognition in open-world settings, it is important to recognize that these benchmarks may not fully capture the complexities and variations present in real-world scenarios. Real-world applications may involve different types of data, varied distributions, and additional challenges that are not fully accounted for in our study.

Furthermore, the scalability of hard sample generation, as used in our fine-tuning experiments, presents a practical limitation. Generating diverse and representative hard positive and negative samples can be computationally expensive and time-consuming. Scaling up the generation process to cover a wide range of positive and negative cases with diverse variations poses a significant challenge and may require more efficient and scalable methods.